\documentclass[10pt, a4paper]{article}

\usepackage[final]{lrec2026} 
\usepackage{booktabs}
\usepackage{multirow}

\title{UniSkill: A Dataset for Matching University Curricula to Professional Competencies}

\name{Nurlan Musazade\textsuperscript{\rm $\dagger$}, Joszef Mezei\textsuperscript{\rm $\dagger$}, Mike Zhang\textsuperscript{{\rm $\ddagger$}{\rm $\diamond$}}}

\address{\textsuperscript{\rm $\dagger$}Åbo Akademi University, Finland \\\textsuperscript{\rm $\ddagger$}University of Copenhagen, Denmark \\
         \textsuperscript{$\diamond$}Pioneer Centre for AI, Denmark \\
         {\tt nurlan.musazade@abo.fi}\\}

\abstract{
Skill extraction and recommendation systems have been studied from recruiter, applicant, and education perspectives. While AI applications in job advertisements have received broad attention, deficiencies in the instructed skills side remain a challenge. In this work, we address the scarcity of publicly available datasets by releasing both manually annotated and synthetic datasets of skills from the European Skills, Competences, Qualifications and Occupations (ESCO) taxonomy and university course pairs and publishing corresponding annotation guidelines. Specifically, we match graduate-level university courses with skills from the Systems Analysts and Management and Organization Analyst ESCO occupation groups at two granularities: course title with a skill, and course sentence with a skill. We train language models on this dataset to serve as a baseline for retrieval and recommendation systems for course-to-skill and skill-to-course matching. We evaluate the models on a portion of the annotated data. Our BERT model achieves 87\% F1-score, showing that course and skill matching is a feasible task.
 \\ \newline  }

\begin{document}

\maketitleabstract

\section{Introduction}

Understanding the skills required for a particular occupation is essential for anyone beginning their career or educational journey. Recent years have seen an upward trend in skill extraction research, particularly from job advertisements \citep{khaouja_survey_2021,senger-etal-2024-deep,zhang2024computational}. Related tasks such as skill clustering \citep{ramazanova_enhancing_2024}, skill recommendation systems \citep{ong_skillrec_2023}, and vacancy ranking or matching for candidates \citep{vanetik_job_2023, kavas-etal-2025-multilingual} have also gained attention.

Despite this growing body of research, \citet{schlippe_skill_2023} emphasizes that effective skill extraction must consider three key stakeholders: employers who demand skills, candidates who supply them, and \emph{educational institutions} that teach them. However, current research has predominantly focused on the employer and candidate perspectives, with limited attention to matching educational resources, such as course syllabi, with real-world job skills. Addressing this ``skill gap'' requires alignment across all three stakeholders \citep{schlippe_skill_2023}, yet the educational perspective remains underexplored.

The few studies conducted in this domain have significant limitations. Most rely on statistical methods~\cite{kitto2020towards} for matching, present only theoretical frameworks~\cite{yu-etal-2021-research}, or focus exclusively on US-centric contexts~\cite{javadian_sabet_course-skill_2024}. Though some semi-open resources exist, they are primarily accessible through commercial contact solutions~\cite{javadian_sabet_course-skill_2024,lightcast_skillabi} rather than being publicly available. To the best of our knowledge, no public dataset currently exists that systematically matches course learning goals to job skills.

To address this gap, we introduce \textbf{UniSkill}, an open-source dataset for aligning course learning goals with standardized job skills. We adopt an education-centric perspective by focusing on the skill supply side, examining learning goals from three course categories: (1) IT courses, (2) Business courses, and (3) Generic courses. These learning goals are mapped to standardized job skills using the European Skills, Competences, Qualifications and Occupations (ESCO) taxonomy~\citep{le2014esco,desmedt-2015-esco}.
With this dataset, we seek to answer two research questions:
\textbf{RQ1.} How effective is the use of individual course sentences for semantic alignment between skills and courses?
\noindent\textbf{RQ2.} How accurate are sentence embedding models in aligning skills and courses based on semantic similarity?

Our results show that using individual course sentences alongside course titles achieves 87\% F1-score with 89\% recall for identifying relevant skills. Furthermore, we investigate the use of synthetic data augmentation on model performance, and show it has a positive effect on model performance.

\paragraph{Contributions.} Our contributions are as follows\footnote{We release data at \url{https://huggingface.co/datasets/nurlanm/UniSkill_dataset} and 

the best-performing model at \url{https://huggingface.co/nurlanm/UniSkill_Bert}}
\begin{itemize}
\itemsep0em
\item \textbf{UniSkill}: The first open-source dataset for aligning course learning goals to standardized occupational skills, consisting of 2,192 annotations: 1,096 sentence-skill pairs and 1,096 course title-skill pairs.
\item Annotation guidelines detailing the process for aligning course learning goals to occupational skills.
\item A synthetic dataset for model training, including prompt engineering guidelines for data generation.
\item Baseline results using bi-encoders and a detailed analysis of false negative predictions.
\end{itemize}


\section{Literature Review}
\label{sec:lit-rev}

\paragraph{Skills and Jobs.}
The analysis of job advertisements plays a crucial role in aligning educational content with labor market needs. A comprehensive survey by \citet{senger-etal-2024-deep} demonstrates that skill extraction granularity ranges from word-level to document-level tasks, ranging from span labeling, binary classification, and coarse-grained categorization. Two primary approaches dominate the field for matching skills to standardized taxonomies: Semantic similarity and extreme multi-label classification \citep{senger-etal-2024-deep, zhang-etal-2024-entity}.

\citet{decorte_extreme_2023} investigate skill extraction using the ESCO taxonomy and publish a synthetic dataset containing skill and job advertisement sentence pairs. Their approach uses bi-encoders with contrastive learning for optimization and cosine similarity for ranking, outperforming previous benchmarks on existing datasets. Their method is straightforward and demonstrates high performance in the job advertisement domain; it provides a foundation for exploring the effectiveness of synthetic data. However, our study diverges by developing a validator model for evaluating ranked semantic similarity results rather than relying solely on retrieval.

Building on this work, \citet{zhang-etal-2024-entity} adopt an entity linking (EL) approach that identifies both explicit span matches and implicit skill mentions. They train their model on the synthetic dataset created by \citet{decorte_extreme_2023}, using actual job advertisements matched to the taxonomy as development and test sets. Their method retrieves the most relevant skill given a sentence by representing skills through both labels and descriptions. However, as the authors acknowledge, this approach and EL methods generally face limitations when evaluation assumes only one correct entity can be linked to a given mention.

\citet{achananuparp_multi-stage_2025} propose a multi-stage framework for occupation classification that also evaluates skill classification performance across different language models. Their framework uses LLMs for inference, retrieval from the ESCO taxonomy, and skill reranking given a job advertisement. Notably, \citet{achananuparp_multi-stage_2025} call for more thorough investigation of ``different choices for top-m retrieval and top-n reranking'', which our validator approach aims to address.



Additional research demonstrates the diversity of ESCO-based approaches. \citet{leon_hierarchical_2024} uses the ESCO taxonomy with embedding models for hierarchical classification of job posting sentences into relevant skill classes. \citet{bocharova_vacancysbert_2023} use ESCO and similarity-based methods for occupation title and skill matching to assist with title normalization. \citet{lavi_consultantbert_2021} fine-tune models using both classification and regression approaches (incorporating cosine similarity) for matching vacancies with job seekers.

\paragraph{Skill and Education.} 

In their comprehensive study of course recommendation systems, \citet{frej_course_2024} define five essential characteristics: alignment with recent skill demands, unsupervised operation, sequential course recommendation, compliance with user goals, and explainability. They identify six key research directions, including dataset scarcity, evaluation metrics that consider job market dynamics, predicting user progress toward objectives, visualization and explainable models, unsupervised competency matching, and dynamic taxonomies incorporating emerging competencies. Our research particularly contributes to the development of unsupervised skill-matching methods.

For skill matching, \citet{frej_course_2024} use a three-step process. First, they use few-shot LLM prompting to extract skills and skill levels from courses. Second, they match three candidate skills from ESCO to each extracted skill using a hybrid approach combining rule-based and cosine similarity methods. The authors note that purely rule-based approaches may miss synonyms, while similarity-based methods risk selecting semantically similar but factually different candidates. Finally, an LLM determines the best match, if any, from the three candidates. Using heuristic evaluation, they find that courses yield an average of seven extracted skills, with approximately four successfully matched to the taxonomy. This study provides valuable methodological guidance for our research.


The most directly relevant work is by \citet{javadian_sabet_course-skill_2024}, who focus on US institutions using the Open Syllabus Project (OSP) and the US-based O*NET taxonomy. While OSP data is publicly available, access requires a research contract. For each syllabus, the authors apply pairwise semantic similarity, defining skill coverage as the highest similarity score between a given skill and any sentence within the course.



Using the same taxonomy with OSP and General Catalog (GC) data, \citet{xu_course_2025} compare TF-IDF, BERT embedding similarity, zero-shot LLMs, and RAG methods. They employ human annotators for benchmark datasets and use LLMs with a calibration model to evaluate extracted skills. Their results show that TF-IDF, BERT, and zero-shot LLM methods achieve approximately 42\% precision at threshold 4 (scoring 4 or 5 out of 5) for the top 10 extracted skills. In contrast, RAG with GPT achieves approximately 82\% precision on OSP data, though performance decreases on GC data. From our perspective, this research highlights the limitations of semantic similarity without additional validation methods. While RAG demonstrates strong performance, it presents scalability challenges due to computational costs. Furthermore, different curriculum document types may yield varying performance compared to standardized open-source datasets like OSP.



\citet{lee_quantifying_2025} study the relationships between courses, occupations, and skills by applying cosine similarity with fixed thresholds. Other alignment measurement approaches exist beyond similarity-based methods. For instance, \citet{ramnarine_using_2025} apply topic modeling and SkillNER to extract skills from curriculum data, revealing general skill gaps across different thematic areas. As noted in the recruitment domain review by \citet{vanetik_job_2023}, while rule-based systems offer straightforward implementation, they may be less robust in handling complex, nuanced, or unusual cases and can generate false positives.

\citet{andonovikj_data-driven_2024} focus on the Slovenian market and a single university, prompting LLaMA models to extract skills and mapping them to ESCO taxonomy using cosine similarity with Sentence-BERT. They retrieve the top ten skills based on similarity scores.

\citet{felby_evaluating_2024} highlight the challenges of temporal dynamics and comparative analysis, calling for automated NLP approaches to study alignment between job advertisements and curricula. They emphasize the need for qualitative analysis involving multiple stakeholders to understand misalignment and context. The authors propose a four-step generic process: (1) focused search for suitable job postings, (2) selecting an analytical framework or taxonomy, (3) coding and categorization, and (4) evaluating alignment. While their study is primarily qualitative, they stress the importance of coding all competencies in both datasets during step three. Our research explicitly addresses the third step by focusing on skill-level coding of curriculum data, while aiming to improve the fourth step of alignment evaluation.

\section{Methodology}

\subsection{Dataset Preparation}

\begin{figure*}[t]
\begin{center}
\includegraphics[width=\linewidth]{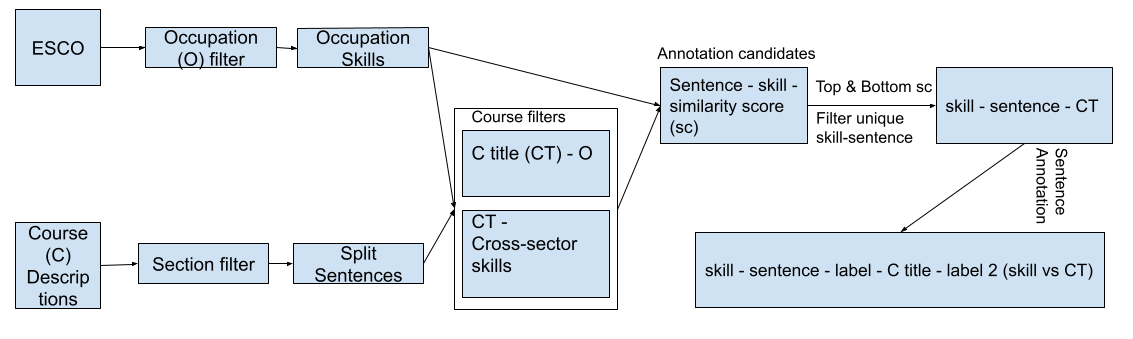}
\caption{The preparation of the original pairs.}
\label{fig.1}
\end{center}
\end{figure*}

\subsubsection{Skill Dataset}

As discussed in the previous section, ESCO and O*NET are the two primary taxonomies used in skill extraction research. Given our European research context and the source of our educational data, we adopt the ESCO taxonomy. ESCO defines approximately 3,000 professions and 14,000 competencies across 28 languages, serving as a unified European skills dictionary.

Due to the large number of occupations and skills in ESCO, we limit our scope to specific professions. Based on the authors' domain knowledge and experience, we select two ESCO occupation groups: ``2511 Systems Analysts'' and ``2421 Management and Organization Analysts''. For each profession within these occupation groups, we retrieve all relevant skills as defined by the ESCO taxonomy. To ensure a manageable dataset size while maintaining diversity of occupations and skills, we work with a representative subset of the occupation-skill pairs from these groups.

\subsubsection{Course Dataset}

To represent instructed skills in educational settings, we use publicly available course descriptions from five Finnish universities. 
We collect graduate-level course data from the past six years (2019-2025). Using section labels in our scraping process, we filter specifically for ``Content'' and ``Learning Objectives'' sections. 
We retain only courses taught in English, filtering based on the ``instruction language'' field when available, or applying automated language detection otherwise. 
Finally, we split each course description into individual sentences. This process yields occupation-skill pairs from the ESCO taxonomy and course-sentence pairs from the university data.
Finding matching course sentences and skills among the vast number of available datapoints is non-trivial; we instead look into pre-annotation matching by using bi-encoders to find possible matching candidates.
Before generating annotation candidates, we conducted preliminary experiments with different models to evaluate our approach and found that we need to combine both semantic similarity and human validation.


\subsubsection{Annotation Candidates}

Given the size of our course dataset, we needed to avoid generating a disproportionate number of true negative pairs. Therefore, we used a pre-trained skill similarity model~\citep{alvperez2025skillsim} to generate candidate pairs for annotation. 
We tried several other models, such as JobBERT-v3~\cite{decorte2025multilingualjobbertcrosslingualjob} and ESCOXLM-R~\cite{zhang-etal-2023-escoxlm}, but found that this pre-trained model worked best.
Following initial observations, we incorporated course titles as contextual information and retained only occupation-relevant and general courses. 
Given the varying lengths of skills, occupations, and titles, this approach enables efficient computation of embeddings and similarity scores on a smaller, more relevant subset of sentences. 
Specifically, we first applied similarity matching between occupation titles and course titles to identify relevant courses. 
For cross-sector skills (e.g., ``teamwork principles'', ``communication''), we generated candidates based on skill-to-course-title similarity to capture pairs from generic courses. 
Finally, for the identified relevant courses, we computed similarity between skills and individual course sentences, selecting both high-similarity (hard) and low-similarity (easy) pairs.

Since annotation was conducted at the skill-sentence level, we removed duplicate rows that arose from cases such as different occupations sharing the same skill or different course years containing identical sentences.

\subsubsection{Annotation Process}

Two authors conducted annotation over five rounds. In rounds 1 through 4, they annotated 50 pairs each of ``hard'' examples (high cosine similarity scores). In round 5, they annotated 50 hard and 50 ``easy'' (semantically least similar) pairs. After each round, annotators discussed disagreements and refined the annotation guidelines accordingly.

In linguistics, metrics for inter-annotator agreement (two annotators) include F1-score, accuracy, and Cohen's kappa (two annotators)~\citep{cohen1960coefficient}. Cohen's kappa is less interpretable than simple percent agreement but accounts for the probability of chance agreement~\citep{mchugh_interrater_2012}. When annotators are well-trained and the likelihood of random guessing is minimal, percent agreement is highly reliable \citep{mchugh_interrater_2012}. Furthermore, acceptable agreement levels should be determined based on domain-specific considerations. In some cases, kappa may underestimate actual agreement due to its underlying assumptions~\citep{mchugh_interrater_2012}. For the 300 course sentence-skill pairs annotated by both annotators, we achieved a Cohen's kappa of approximately 0.45, indicating moderate agreement. The F1-score for class ``0'' (non-match) was 0.76 and for class ``1'' (match) was 0.69, with an overall F1-score of 0.73, which shows a reasonable agreement. Since a single course contains multiple semantically similar sentences, the likelihood of missing a relevant course in practice is low. Therefore, we consider the current agreement level reliable for proceeding with annotation. The main annotator continued the annotation process independently.

Table~\ref{tab:data-split} shows the class distribution across splits. In total, we annotated approximately 2,200 pairs of skill-course\_sentence and skill-course\_title combinations, covering 406 unique skills, 787 sentences, and 561 titles. Approximately 70\% (750 pairs) are hard pairs with high semantic similarity. Of these, 300 pairs (50\% hard) come from the ``Management and Organization Analyst'' occupation group, while 800 pairs come from the ``Systems Analysts'' occupation group.

\begin{table}[t]
\centering
\small
\begin{tabular}{lccc}
\toprule
\textbf{Label} & \textbf{Train} & \textbf{Validation} & \textbf{Test} \\
\midrule
\multicolumn{4}{l}{\textit{Non-match (Class 0)}} \\
\quad Sentence-Skill & 434 & 36 & 171 \\
\quad Title-Skill & 331 & 14 & 145 \\
\midrule
\multicolumn{4}{l}{\textit{Match (Class 1)}} \\
\quad Sentence-Skill & 261 & 66 & 130 \\
\quad Title-Skill & 364 & 88 & 156 \\
\bottomrule
\end{tabular}
\caption{\textbf{Data Distribution.} Class distribution across training, validation, and test splits for course sentence-skill, course title-skill, and combined pairs.}
\label{tab:data-split}
\end{table}

\subsection{Synthetic Data Generation}

After annotating the original course-skill pairs, we first tested whether incorporating existing synthetic data could improve performance. Specifically, we experimented with the synthetic job advertisement sentence and ESCO skill pairs created by \citet{decorte_extreme_2023}. 
Due to their large dataset size, we filtered only skills that appeared in our manually annotated dataset. We assigned a label of ``1'' (match) to all pairs containing these skills. 
For negative pairs, we randomly selected sentences from the dataset that did not belong to the rows associated with each skill. This process generated approximately 7,200 pairs with balanced positive and negative labels.

Initial testing with a BERT model revealed a performance drop compared to using only our annotated dataset. Even when training with only positive pairs (an admittedly unreliable approach for creating balanced training data), we observed decreased performance. These results demonstrate that while the task and skills remain the same, differences in textual data and context are critical. This finding underscores the necessity of creating a task-specific training dataset tailored to educational content rather than job advertisements.

Following the methodology of \citet{decorte_extreme_2023}, we adapted their prompting approach to generate course sentences instead of job advertisement sentences. 
Due to computational costs and the iterative nature of prompt engineering, \citet{decorte_extreme_2023} used the gpt-3.5-turbo-0301 model, achieving 94\% accuracy while covering 99.5\% of ESCO skills. More recently, \citet{achananuparp_multi-stage_2025} studied LLM capacity for generating occupation titles from different taxonomies, finding that GPT-4o significantly outperforms gpt-3.5-turbo in most cases. While we used the user prompt from \citet{decorte_extreme_2023} as our starting point, we selected GPT-4o for synthetic data generation based on its superior performance in similar domain tasks and its more recent release.

Given the importance of implicit versus explicit skill mentions discussed in the literature \citep{zhang-etal-2024-entity}, we prompted the model to generate two sentences per skill: one with implicit mention and another with explicit mention. Since the task is straightforward without complex formatting requirements, we began with zero-shot settings. We first tested the model on 10 skills (generating 20 sentences) to evaluate performance before proceeding at scale. To ensure representativeness, we manually selected 5 tool-based skills and 5 soft skills or concepts (e.g., cybersecurity).

Following \citet{decorte_extreme_2023}, we used a system prompt with additional instructions tailored to our sentence types. We adopted their approach of defining a clear user input format structure consisting of the number of sentences required, the skill name, and the skill definition. We retrieved official skill definitions directly from the ESCO taxonomy. After multiple refinement iterations (without API calls), we finalized our prompt. We will release the prompts upon paper acceptance. We manually reviewed the 20 test sentences to evaluate prompt effectiveness, finding that the model performed as instructed for all pairs. We then executed the script for all 406 unique skills in our manually annotated dataset, generating approximately 800 positive pairs.

We followed the same validation process, including manual review of 20 sentences, using a different system prompt for generating irrelevant course sentences. Finally, for each skill-sentence pair, we created both matching and non-matching course titles. Through this process, we refined six different prompts: two for positive and negative sentence pairs, and four for course titles (matching and non-matching for both positive and negative sentences). This resulted in approximately 3,200 skill-sentence-course title triplets with equal distribution across four combinations: (1) relevant sentence with relevant course title, (2) relevant sentence with irrelevant course title, (3) irrelevant sentence with relevant course title, and (4) irrelevant sentence with irrelevant course title. We show desciptive statistics of the final dataset in Appendix~\ref{app:descriptive} and additionally a couple of examples in Appendix~\ref{app:samples}.

\subsection{Models, Training, and Evaluation}

\subsubsection{Combined Approach and Data Splits}

Our approach combines course metadata and content by creating pairs of skills with both course titles and course sentences, which we refer to as the Combined Approach (CA). However, we initially test models with titles and sentences separately for two reasons: first, to compare performance against the CA; second, to use separate predictions as potential explanations when analyzing incorrect predictions from the combined model.

We create our data splits by randomly shuffling the annotated dataset and selecting 300 pairs as the test set, 700 pairs for training, and reserving 100 hard pairs (high similarity scores) for additional validation. We use synthetic data only during training to ensure evaluation is performed exclusively on real annotated data.

For the Combined Approach labeling strategy, we assign a positive label only when both the skill-course title pair and the skill-course sentence pair are annotated as matches. All other combinations receive a negative label. This conservative labeling ensures that a course is considered relevant only when both the title and content align with the skill.

\subsubsection{Model Input Formatting}

We experiment with model performance using different input formatting strategies: with and without synthetic data, and with and without special tokens (i.e.\texttt{[TITLE]}, \texttt{[SENTENCE]}, \texttt{[SKILL]}, \texttt{[DESCRIPTION]}) to delineate different text components, testing whether using explicit markers to separate course titles, course sentences, skill names, and skill descriptions improves model performance. This approach follows prior work such as \citet{bocharova_vacancysbert_2023}, who use special tokens to help models distinguish between different information types in the input sequence.

\subsubsection{Model Selection and Training}

We begin by testing a BERT model~\cite{devlin-etal-2019-bert} as our baseline to establish initial performance benchmarks. After determining the optimal hyperparameter settings through baseline experiments (comparing synthetic data inclusion, input formatting strategies, and the CA versus separate models), all on the development set, we proceed to train and evaluate domain-specific models.

To ensure model diversity across training datasets, languages, and training objectives, we select five domain-relevant models. 
The labor\_space model \citep{kim_labor_2024} is explicitly trained on labor market data. 
The GBERT model \citep{gnehm-etal-2022-evaluation} is a German BERT variant with potential multilingual capabilities. 
The ESCOXLM-R model \citep{zhang-etal-2023-escoxlm} is a multilingual model pre-trained on ESCO taxonomy data. 
The ESCOXLM-R\_ENG model \citep{musazade-etal-2025-cross} is a fine-tuned version of ESCOXLM-R specifically adapted for binary sentence-level skill identification in English. Finally, the me5-base-course-skill model \citep{sheikh_learning_2024} is a multilingual model fine-tuned on German ESCO skills and courses.

For all models, we maintain consistent training datasets and hyperparameters to ensure fair comparison (Appendix~\ref{app:hyperparams}). This allows us to isolate the impact of model architecture and pre-training objectives on performance for our specific task of skill-course alignment.

\begin{table*}[t]
    \centering
    \small
    \begin{tabular}{llcccccc}
\toprule
\textbf{ST} & \textbf{Approach} & \textbf{Class} & \textbf{F1} & \textbf{Prec.} & \textbf{Rec.} & \textbf{Avg F1} & \textbf{Acc.} \\
\midrule
\multirow{6}{*}{Without} 
& \multirow{2}{*}{Course sentence-Skill} & 0 & 0.87 & 0.84 & \textbf{0.91} & \multirow{2}{*}{0.84} & \multirow{2}{*}{0.85} \\
& & 1 & 0.81 & 0.86 & 0.77 & & \\
\cmidrule{2-8}
& \multirow{2}{*}{Course title-Skill} & 0 & 0.87 & 0.83 & 0.90 & \multirow{2}{*}{\textbf{0.87}} & \multirow{2}{*}{\textbf{0.87}} \\
& & 1 & \textbf{0.87} & \textbf{0.90} & 0.83 & & \\
\cmidrule{2-8}
& \multirow{2}{*}{Combined} & 0 & --- & --- & --- & \multirow{2}{*}{---} & \multirow{2}{*}{---} \\
& & 1 & --- & --- & --- & & \\
\midrule
\multirow{6}{*}{With} 
& \multirow{2}{*}{Course sentence-Skill} & 0 & 0.87 & 0.85 & 0.89 & \multirow{2}{*}{0.85} & \multirow{2}{*}{0.85} \\
& & 1 & 0.82 & 0.85 & 0.79 & & \\
\cmidrule{2-8}
& \multirow{2}{*}{Course title-Skill} & 0 & 0.86 & \textbf{0.96} & 0.79 & \multirow{2}{*}{\textbf{0.88}} & \multirow{2}{*}{\textbf{0.88}} \\
& & 1 & \textbf{0.89} & 0.83 & \textbf{0.97} & & \\
\cmidrule{2-8}
& \multirow{2}{*}{Combined} & 0 & \textbf{0.89} & 0.90 & 0.88 & \multirow{2}{*}{0.85} & \multirow{2}{*}{0.86} \\
& & 1 & 0.81 & 0.79 & 0.82 & & \\
\bottomrule
\end{tabular}
\caption{\textbf{Separate Model Results.} Performance comparison of separate sentence-only, title-only, and combined predictions with and without input labels. All models were trained with synthetic data. We either use Special Tokens (ST) or not. Class 0 = Non-match, Class 1 = Match.}
\label{tab:separate-results}
\end{table*}

\section{Experimental Results}

\subsection{Separate Course Title and Course Sentence Matching Models}

We initially evaluated the BERT model separately for skill-course title pairs and skill-sentence pairs. As shown in Table~\ref{tab:separate-results}, all metrics demonstrate the embedding models' capability to perform this task. The course title-skill classification achieves higher performance with an average F1-score of 0.88, which is 0.03 points higher than the course sentence-skill classification task. Notably, we observe lower recall for positive matches in the sentence classification task, falling below 80\%.

We experiment with two input formatting strategies: with and without special tokens to mark different text segments, as described in the methodology section. While differences are observable, the variations are relatively small. This is expected given that each input contains only a single text segment (either sentence or title) paired with a skill. However, we also apply labels to distinguish between skill names and skill descriptions. The results show that labeling has a more pronounced effect on the course title approach, where recall for positive matches is higher with labels, but precision is lower. At the same time, the pattern reverses for negative matches.

To compare the separate classification approach with the Combined Approach (CA) presented in Table~\ref{tab:combined-results}, we merge predictions from the sentence-only and title-only models for each course. We assign a positive label only when both models predict a match, and a negative label otherwise. As shown in Table~\ref{tab:separate-results} in the ``combined predictions'' column, this provides a more transparent comparison with the CA than simply averaging metrics. The combined separate models achieve an F1 score of approximately 85\%, with higher performance on negative matches than positive matches.

\subsection{Combined Approach (CA)}

In our second approach, shown in Table~\ref{tab:combined-results}, we use a single model that takes both title and sentence as combined input. The results show slightly higher overall performance than the separate models approach, with an F1 score improvement of approximately 0.015. However, the most notable difference is a 7\% increase in recall for positive matches in the CA. We prioritize higher recall for the relevant class to minimize false negatives and avoid missing relevant courses. Additionally, in both positive and negative classes, we observe slightly higher performance when using labeled input formatting. Therefore, we select the combined input of course title and course sentence with labels as the optimal approach for this task.

To analyze the contribution of synthetic data, we test the selected approach using only manually annotated training data. As shown in Table~\ref{tab:combined-results}, the results demonstrate a substantial positive effect of synthetic data on model learning. For positive matches, models trained with synthetic data achieve 6\% and 8\% higher recall than models trained without synthetic data in labeled and unlabeled settings, respectively. Even in the annotated-only dataset setting, we observe the positive impact of labeling on recall for positive matches. Due to inconsistent results, likely attributable to the limited dataset size, we train each model three times without synthetic data and report the average performance.

To further evaluate performance on challenging cases, we test the trained CA models (with and without labels) on 100 additional hard pairs with high semantic similarity. As shown in Table~\ref{tab:hard-results}, despite the dataset consisting exclusively of difficult pairs, we achieve an overall accuracy of 74\%. Most importantly, recall for relevant pairs reaches approximately 79\%, which is only 10\% lower than performance on the mixed test set shown in Table~\ref{tab:combined-results}. Once again, the positive impact of input labeling is evident.

\begin{table}[t]
\centering
\small
\begin{tabular}{llcccc}
\toprule
\textbf{Synthetic} & \textbf{ST} & \textbf{Class} & \textbf{F1} & \textbf{Prec.} & \textbf{Rec.} \\
\midrule
\multirow{2}{*}{Yes} & \multirow{2}{*}{No} 
& 0 & 0.89 & 0.92 & 0.87 \\
& & 1 & 0.82 & 0.78 & 0.86 \\
\cmidrule{3-6}
& & Avg & 0.86 & 0.85 & 0.87 \\
\midrule
\multirow{2}{*}{Yes} & \multirow{2}{*}{Yes} 
& 0 & \textbf{0.90} & \textbf{0.93} & 0.87 \\
& & 1 & \textbf{0.83} & \textbf{0.79} & \textbf{0.89} \\
\cmidrule{3-6}
& & Avg & \textbf{0.87} & \textbf{0.86} & \textbf{0.88} \\
\midrule
\multirow{2}{*}{No} & \multirow{2}{*}{Yes} 
& 0 & 0.86 & 0.90 & 0.83 \\
& & 1 & 0.78 & 0.74 & 0.83 \\
\cmidrule{3-6}
& & Avg & 0.82 & 0.82 & 0.83 \\
\midrule
\multirow{2}{*}{No} & \multirow{2}{*}{No} 
& 0 & 0.87 & 0.89 & 0.85 \\
& & 1 & 0.78 & 0.76 & 0.81 \\
\cmidrule{3-6}
& & Avg & 0.83 & 0.83 & 0.83 \\
\bottomrule
\end{tabular}
\caption{Combined Approach performance with different training configurations. We either use Special Tokens (ST) or not. Class 0 = Non-match, Class 1 = Match. Best performance (bold averages) achieved with synthetic data and input labels.}
\label{tab:combined-results}
\end{table}

\begin{table}[t]
\centering
\small
\begin{tabular}{llcccc}
\toprule
\textbf{ST} & \textbf{Class} & \textbf{F1} & \textbf{Prec.} & \textbf{Rec.} & \textbf{Acc.} \\
\midrule
\multirow{3}{*}{No} 
& 0 & 0.63 & 0.60 & 0.67 & \multirow{3}{*}{0.68} \\
& 1 & 0.72 & 0.75 & 0.69 & \\
\cmidrule{2-5}
& Avg & 0.68 & 0.68 & 0.68 & \\
\midrule
\multirow{3}{*}{Yes} 
& 0 & 0.68 & 0.69 & 0.67 & \multirow{3}{*}{\textbf{0.74}} \\
& 1 & \textbf{0.78} & \textbf{0.77} & \textbf{0.79} & \\
\cmidrule{2-5}
& Avg & \textbf{0.73} & \textbf{0.73} & \textbf{0.73} & \\
\bottomrule
\end{tabular}
\caption{Combined Approach performance on 100 hard validation pairs with high semantic similarity. We either use Special Tokens (ST) or not. Class 0 = Non-match, Class 1 = Match. Both models trained with synthetic data.}
\label{tab:hard-results}
\end{table}

\subsection{Domain-Specific Models}

Finally, we evaluate domain-specific models using the highest performing configuration: labeled input, synthetic data, and the Combined Approach (CA). Surprisingly, while the ESCOXLM-R\_ENG and jobGBERT models achieve higher precision for positive matches, all domain-specific models fall behind the baseline BERT model in terms of recall. Among the domain models, esco-xlm-roberta-large achieves the highest recall for positive matches at 82\%, which is still 7\% lower than the BERT baseline. Overall, ESCOXLM-R\_ENG demonstrates the highest accuracy at 86\%, while both ESCOXLM-R\_ENG and esco-xlm-roberta-large achieve the highest F1 scores at approximately 84\%.

\begin{table}[t]
\centering
\scriptsize
\begin{tabular}{llcccc}
\toprule
\textbf{Model} & \textbf{Class} & \textbf{F1} & \textbf{Prec.} & \textbf{Rec.} & \textbf{Acc.} \\
\midrule
\multirow{3}{*}{\shortstack[l]{ESCOXLM-R\_ENG}} 
& 0 & \textbf{0.89} & 0.87 & \textbf{0.91} & \multirow{3}{*}{\textbf{0.86}} \\
& 1 & 0.79 & 0.83 & 0.76 & \\
\cmidrule{2-5}
& Avg & \textbf{0.84} & \textbf{0.85} & \textbf{0.84} & \\
\midrule
\multirow{3}{*}{\shortstack[l]{ESCOXLM-R}} 
& 0 & 0.88 & \textbf{0.90} & 0.85 & \multirow{3}{*}{0.84} \\
& 1 & 0.79 & 0.76 & 0.82 & \\
\cmidrule{2-5}
& Avg & \textbf{0.84} & 0.83 & \textbf{0.84} & \\
\midrule
\multirow{3}{*}{labor\_space} 
& 0 & 0.86 & 0.88 & 0.83 & \multirow{3}{*}{0.82} \\
& 1 & 0.76 & 0.73 & 0.80 & \\
\cmidrule{2-5}
& Avg & 0.81 & 0.81 & 0.82 & \\
\midrule
\multirow{3}{*}{jobGBERT} 
& 0 & 0.88 & 0.83 & \textbf{0.93} & \multirow{3}{*}{0.83} \\
& 1 & 0.74 & 0.84 & 0.66 & \\
\cmidrule{2-5}
& Avg & 0.81 & 0.84 & 0.80 & \\
\midrule
\multirow{3}{*}{\shortstack[l]{me5-base-course-skill}} 
& 0 & 0.88 & 0.88 & 0.88 & \multirow{3}{*}{0.85} \\
& 1 & 0.79 & 0.79 & 0.79 & \\
\cmidrule{2-5}
& Avg & \textbf{0.84} & \textbf{0.84} & \textbf{0.84} & \\
\bottomrule
\end{tabular}
\caption{Domain-specific model performance on the test set (300 pairs). All models trained with synthetic data, input labels, and the Combined Approach. Class 0 = Non-match, Class 1 = Match.}
\label{tab:domain-models}
\end{table}

\section{Analysis of False Negatives}

As a final step, we analyze incorrect predictions to identify common patterns in false negatives, which are particularly important for this task. Our findings reveal that for the 12 false negatives (FNs) produced by the Combined Approach, the separate title model predicted all pairs as relevant. In contrast, the separate sentence model predicted only three as relevant. Recall that in the Combined Approach, we label a pair as appropriate only when both title and sentence align with the skill. In other words, based on our analysis of FNs from the CA model (BERT with 0.89 recall in Table~\ref{tab:combined-results}), the sentence component was the primary reason for assigning a negative prediction, even though the model likely treated the title as relevant.

All 12 FNs come from pairs we categorized as hard to annotate due to high semantic similarity. From the skill perspective, we observe equal distribution across three categories: data and databases, ICT and software, and general concepts such as ``engineering processes'' and ``mentor individuals.'' Regarding sentence context, we observe that in most cases, we adopted a broader interpretation of relevance during annotation. For instance, we considered the skill ``create data sets'' as being taught in a sentence discussing trend discovery and data mining. Similarly, we matched ``cloud technologies'' to the sentence ``in the cloud context identify and analyze the strengths and domains of different programming paradigms and languages,'' ``forecast future ICT network needs'' to ``Practical examples of commonly used network architectures, protocols and algorithms in the Internet,'' ``design database scheme'' to ``Discover trends in analytical data stores using the data mining techniques,'' and ``deploy ICT systems'' to ``Design principles for distributed applications and services.'' In three cases, both the sentence and title would have been predicted as relevant if evaluated separately, resulting in true positives.

To further investigate this pattern, we examine the 13 FNs from the additional validation set shown in Table~\ref{tab:hard-results}. When predictions are made separately, 11 out of 13 titles and 9 out of 13 sentences would be classified as relevant. However, when we combine the separate predictions using the same logic as the CA (requiring both to be positive), we obtain 12 FNs. Notably, the pairs misclassified by the separate approach differ from those misclassified by the combined approach. 

Overall, this review demonstrates that FNs are indeed complex pairs where relevance is ambiguous. With a stricter interpretation during annotation, these pairs could reasonably be labeled as non-relevant. This analysis suggests that while the model achieves strong overall performance, access to more detailed learning objectives within course descriptions would ensure safer predictions and higher performance by reducing ambiguity in skill-course alignment.

\section{Conclusion}

Our results demonstrate the effectiveness of using language models as validators to determine whether a course is relevant for acquiring a specific skill. The LLM achieves at least human-level agreement in this complex and challenging classification task. The study illustrates from multiple perspectives the critical importance of context in improving language model performance. First, we show that incorporating course titles is essential for context awareness. Second, using a single validator model with combined sentence and course title as input achieves the highest performance. Finally, labeling both input and output components positively affects the classification outcomes of BERT models.

Furthermore, aligned with previous studies on job advertisements, our research demonstrates that careful prompting can achieve high-quality synthetic data generation with relatively modest effort. While manually annotated data provides a reliable evaluation set, synthetic data substantially improves model performance when predicting alignment in real course-skill pairs. However, we find that synthetic data generated for skill extraction from job advertisements does not necessarily improve model performance for skill matching in educational course contexts, even though the underlying task remains the same.

Future research should investigate the diverse variables involved in generating synthetic course descriptions. The presented dataset establishes appropriate contextual distinctions by specifying, for example, whether skills are mentioned explicitly or implicitly, and whether they represent specific tools or broader competencies. Additionally, future work should evaluate the validator approach within the complete pipeline of course recommendation systems to assess its practical utility in real-world educational settings.

\section*{Limitations}

This study focuses on a limited set of occupational groups. Future research could examine the generalizability of the findings to other occupations, study levels, and languages. Future work could investigate how characteristics of course descriptions, such as institution- or country-specific phrasing, or sections, affect model performance. Exploring incorporating emerging skills that may not yet be defined in ESCO would improve coverage.

\section*{Ethical Considerations}
The dataset was collected from publicly available university course descriptions, and we confirmed with a representative from one of the universities included in the data collection that the use and publication of such data for research purposes is acceptable.

Systems built on or deploying our model and dataset should carefully communicate to end users the definition of relevance and the annotation guidelines applied in this work. Misinterpretation of the models or systems' outputs, or assuming that they represent objective and universal skill-course relationships, could lead to misleading educational or policy conclusions. Transparency on the underlying matching logic and limitations is essential for responsible use. Moreover, even high performance achieved does not guaranty inclusion of all relevant courses, creating the risk of enlarging the educational gap if deployed incautiously. 

\section*{Acknowledgments}
NM received funding from the GUSTAF PACKALÉNS MINDEFOND supporting his visit to Aalborg University, Denmark. MZ was supported by the research grant (VIL57392) from VILLUM FONDEN and also received funding from the Danish Government to Danish Foundation Models (4378-00001B).

\section{Bibliographical References}\label{sec:reference}
\bibliographystyle{lrec2026-natbib}
\bibliography{lrec2026-example,anthology-1,anthology-2}


\clearpage
\appendix

\section{Hyperparameters}\label{app:hyperparams}
We show our hyperparameter settings in Table 6.

\section{Dataset Description}\label{app:descriptive}
We show several descriptive statistics in Table 7, 8, and 9.

\section{Dataset Pair Samples}\label{app:samples}
We show a couple of paired examples in Table 10.

\begin{table*}[t]
\centering
\begin{tabular}{ll}
\toprule
Parameter & Value \\
\midrule
Model architecture & AutoModelForSequenceClassification \\
Learning rate & $2 \times 10^{-5}$ \\
Optimizer & AdamW \\
Training batch size & 16 (shuffle=True) \\
Evaluation batch size & 32 \\
Epochs & 7 \\
Max sequence length & 128 \\
Truncation & longest\_first \\
Padding & max\_length \\
Random seed & 42 \\
Number of labels & 2 \\
torch.backends.cudnn.deterministic & True \\
torch.backends.cudnn.benchmark & False \\
\bottomrule
\end{tabular}
\label{hyperparams}
\caption{Training hyperparameters.}
\end{table*}

\begin{table}[t]
\centering
\begin{tabular}{l r}
\toprule
\textbf{Statistics} & \textbf{Value}\\
\midrule
Total pairs & 301 \\
Positive pairs & 108 \\
Negative pairs & 193 \\
\midrule
Avg. sentence length & 19.32 \\
Median sentence length & 12 \\
Std. sentence length & 25.69 \\
\midrule
Avg. title length & 4.4 \\
Median title length & 4 \\
Std. title length & 2.01 \\
\midrule
Unique ESCO skills & 216 \\
Unique courses & 235 \\
Avg. skills per course (Matches) & 1.18 \\
Max. skills per course (Matches) & 4 \\
\bottomrule
\end{tabular}
\caption{Descriptive statistics of the evaluation dataset.}
\label{tab:eval_stats}
\end{table}

\begin{table}[t]
\centering
\begin{tabular}{l r}
\toprule
\textbf{Statistics} & \textbf{Value}\\
\midrule
Total pairs & 102 \\
Positive pairs & 60 \\
Negative pairs & 42 \\
\midrule
Avg. sentence length & 18.07 \\
Median sentence length & 12.5 \\
Std. sentence length & 19.14 \\
\midrule
Avg. title length & 4.22 \\
Median title length & 4 \\
Std. title length & 1.44 \\
\midrule
Unique ESCO skills & 86 \\
Unique courses & 78 \\
Avg. skills per course (Matches) & 1.28 \\
Max. skills per course (Matches) & 3 \\
\bottomrule
\end{tabular}
\caption{Descriptive statistics of the additional validation dataset.}
\label{tab:val_stats}
\end{table}

\begin{table}[t]
\centering
\begin{tabular}{l r}
\toprule
\multicolumn{2}{l}{\textbf{Original pairs}} \\
\midrule
Total pairs & 695 \\
Positive pairs & 230 \\
Negative pairs & 465 \\
\midrule
Avg. sentence length & 17.52 \\
Median sentence length & 11 \\
Std. sentence length & 25.62 \\
\midrule
Avg. title length & 4.26 \\
Median title length & 4 \\
Std. title length & 2.22 \\
\midrule
Unique ESCO skills & 351 \\
Unique courses & 424 \\
Avg. skills per course (Matches) & 1.40 \\
Max. skills per course (Matches) & 6 \\
\midrule
\multicolumn{2}{l}{\textbf{Synthetic pairs}} \\
\midrule
Total pairs & 3212 \\
Positive pairs & 802 \\
Negative pairs & 2410 \\
\midrule
Avg. sentence length & 12.42 \\
Median sentence length & 12 \\
Std. sentence length & 3.76 \\
\midrule
Avg. title length & 5.57 \\
Median title length & 5 \\
Std. title length & 1.54 \\
\midrule
Unique ESCO skills & 401 \\
Unique courses & 2779 \\
Avg. skills per course (Matches) & 1.01 \\
Max. skills per course (Matches) & 3 \\
\bottomrule
\end{tabular}
\caption{Descriptive statistics of the training dataset.}
\label{tab:train_stats}
\end{table}

\begin{table*}[t]
\centering
\small
\begin{tabular}{p{4cm} p{7cm} p{2cm} c}
\toprule
Course Title & Course Sentence & ESCO Skill & Label \\
\midrule
Food Metabolomics and Multivariate Analysis & Introduction to different statistical methods (e.g. &
statistics & 0 \\
\addlinespace
Artificial Intelligence for Teachers Open University Studies & Jupyter notebooks can be edited and run in a standard browser, using freely available tools such as Google Collaboratory. &
TypeScript & 0 \\
\addlinespace
Statistical Thinking and Performing Basic Analyses Using JMP &
The objective is to understand the basic concepts and methods of descriptive and inferential statistical analysis. &
statistics & 1 \\
\addlinespace
Cybersecurity D &
Cyber security concepts. &
attack vectors & 1 \\
\addlinespace
Project Management and Consulting Practice &
The project management part deals with advanced issues of project work. &
Agile project management & 1 \\
\bottomrule
\end{tabular}
\caption{Example course - skill pairs from the real (evaluation) dataset.}
\label{tab:examples}
\end{table*}

\end{document}